\documentclass[letterpaper]{article} % DO NOT CHANGE THIS
\usepackage{aaai2027}  % DO NOT CHANGE THIS
\nocopyright
\usepackage[hyphens]{url}  % DO NOT CHANGE THIS
\usepackage{graphicx} % DO NOT CHANGE THIS
\usepackage{natbib}  % DO NOT CHANGE THIS AND DO NOT ADD ANY OPTIONS TO IT
\usepackage{caption} % DO NOT CHANGE THIS AND DO NOT ADD ANY OPTIONS TO IT
\usepackage{algorithm}
\usepackage{algorithmic}
\usepackage{amsmath,amsfonts,amssymb}
\usepackage{booktabs}
\usepackage{multirow}
\usepackage{subfig}
\usepackage{booktabs}
\usepackage{multirow}
\usepackage[table]{xcolor}

\definecolor{bestred}{RGB}{190,35,35}
\definecolor{secondblue}{RGB}{35,35,190}
\definecolor{thirdgreen}{RGB}{0,125,80}

\newcommand{\best}[1]{\textbf{\textcolor{bestred}{#1}}}

\newcommand{\second}[1]{\textbf{\textcolor{thirdgreen}{#1}}}
\usepackage{newfloat}
\usepackage{listings}
\DeclareCaptionStyle{ruled}{labelfont=normalfont,labelsep=colon,strut=off} % DO NOT CHANGE THIS
\floatstyle{ruled}
\newfloat{listing}{tb}{lst}{}
\floatname{listing}{Listing}

\usepackage{booktabs}

\title{SpecUltra: A Unified Foundation Model for Heterogeneous Spectral Images}
\author{
    Weili Kong\textsuperscript{\rm 1}\thanks{Main contribution},
    Baisen Liu\textsuperscript{\rm 1}\textsuperscript{\rm 2}\corresponding,
    Xiaojun Bi\textsuperscript{\rm 1}\textsuperscript{\rm 3},
    Jiaming Pei\textsuperscript{\rm 4}\\
    {kkweil, liubaisen, bixiaojun}@hrbeu.edu.cn, 
    jpei0906@uni.sydney.edu.au
}
\affiliations{
    \textsuperscript{\rm 1}School of Information and Communication Engineering, Harbin Engineering University, Harbin 150001, China\\
    \textsuperscript{\rm 2}School of Electronic and Information Engineering, Heilongjiang Institute of Technology, Harbin 150050, China\\
    \textsuperscript{\rm 3}School of Information Engineering, Minzu University of China, Beijing 100081, China\\
    \textsuperscript{\rm 4}School of Computer Science, The University of Sydney, Sydney, NSW 2006, Australia 
}

\begin{document}

\maketitle

\begin{abstract}
Spectral imagery, including hyperspectral image (HSI) and multispectral image (MSI), exhibits substantial heterogeneity in band numbers, wavelengths, spectral resolutions, hindering unified modeling and cross-sensor transfer. 
From the perspective of imaging physics, these heterogeneous observations can be regarded as discrete measurements of an underlying continuous spectrum acquired through different sampling processes. Motivated by this perspective, we propose \textit{\textbf{SpecUltra}}, a unified foundation model for heterogeneous spectral images. Specifically, SpecUltra employs 3D convolutional tokenization and wavelength-aware 3D RoPE to encode relative wavelength and spatial relationships, enabling fine-grained spatial-spectral modeling within a shared coordinate system. We further construct \textit{\textbf{MultiHyper-200K}}, a large-scale multi-sensor HSI dataset, and pretrain SpecUltra using auxiliary-token-assisted mask image modeling (MIM), where a level token indicates the radiometric product type and register tokens absorb sensor- and acquisition-dependent variations.
To bridge the spectral-granularity gap between HSI and MSI, we introduce Spectral Kernel Transformation (SKT), which transfers pretrained knowledge in parameter space by constructing MSI embedding kernels from fine-grained HSI-pretrained kernel slices. This design preserves pretrained spectral priors while maintaining compatibility with the pretrained backbone.
Experiments on HSI and MSI classification and segmentation benchmarks demonstrate state-of-the-art or competitive performance across disparate sensors and downstream tasks.
\end{abstract}

\begin{links}
    \link{Code}{https://github.com/kkweil/SpecUltra}
\end{links}

\section{Introduction}
Spectral imagery jointly captures the spatial structures of ground objects and their wavelength-dependent reflectance characteristics, providing rich information for material identification, target recognition, and land-cover classification~\cite{ghamisi2017advances, ayala2023spectral}. Unlike RGB images with fixed channel definitions, spectral images exhibit substantial heterogeneity in band configurations. HSI densely samples the spectrum using tens to hundreds of contiguous narrow bands, whereas MSI records integrated responses over only a few discrete and relatively broad spectral intervals~\cite{amigo2019hyperspectral}. Moreover, even within the same modality, different sensor designs can lead to substantial variations in band number, central wavelength, bandwidth, and spectral response functions. This heterogeneity poses substantial challenges to building a unified spectral foundation model for spectral images.

\begin{figure}[t]
\centering
\includegraphics[width=\columnwidth]{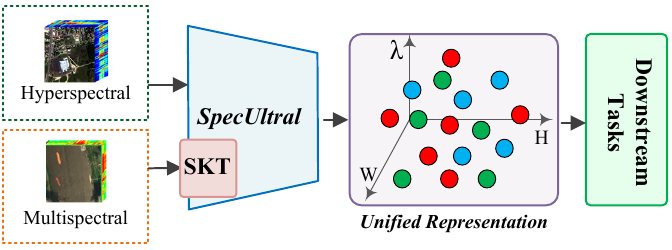} 
\caption{SpecUltra unifies heterogeneous HSI and MSI in a shared representation space and is subsequently finetuned for downstream tasks.}
\label{fig1}
\end{figure}

A natural strategy is to embed each spectral channel independently, thereby preserving fine-grained spectral information. However, for HSI with hundreds of densely sampled bands, this approach results in prohibitively long token sequences and incurs substantial computational overhead. Spectral group embedding~\cite{hong2022spectralformer} provides a more efficient alternative by jointly encoding neighboring bands and exploiting the local spectral patterns of HSI. However, this design is less suitable for MSI, whose bands are sparse and relatively broad. Grouping adjacent MSI bands may prematurely mix distinct spectral semantics.
Consequently, existing spectral foundation models typically specialize in either HSI or MSI~\cite{hong2026foundation}, with their input embeddings and positional representations tied to specific band configurations~\cite{hong2026hyperspectral}. Although sensor-agnostic methods such as SMARTIES~\cite{sumbul2025smarties} and CARL~\cite{baumann2026carl} employ wavelength-aware projections or adaptive spectral aggregation, they often compress spectral information early or rely on sensor-dependent mappings, limiting fine-grained spatial-spectral modeling.

From the perspective of spectral imaging physics, HSI and MSI share a common observation process. Both can be regarded as discrete measurements of an underlying continuous surface spectrum, differing primarily in their spectral sampling densities and response functions~\cite{yokoya2017hyperspectral}. This shared physical origin provides a principled basis for unified modeling and suggests that knowledge learned from densely sampled HSI can be transferred to coarsely sampled MSI by adapting pretrained representations and parameters to different spectral granularities. Accordingly, two fundamental questions arise: \textit{1) How can} heterogeneous spectral observations be modeled with fine-grained spatial-spectral structures and consistent physical semantics? and \textit{2) How can} pretrained knowledge be effectively transferred across substantially different spectral granularities?

To this end, we propose SpecUltra, a unified foundation model for heterogeneous spectral remote sensing imagery. SpecUltra learns fine-grained spatial-spectral representations from multi-sensor HSI data and transfers them to diverse MSI configurations through parameter-space adaptation.
Specifically, a 3D convolutional embedding layer partitions each input into spatial-spectral tokens. Wavelength-aware 3D RoPE then encodes relative physical relationships along the wavelength and spatial dimensions, enabling heterogeneous observations to be modeled in a unified coordinate system. SpecUltra is pretrained on MultiHyper-200K, which contains approximately 200,000 HSI samples from six sensors, using auxiliary-token-assisted masked image modeling. A level token indicates the radiometric product type, while register tokens capture sensor- and acquisition-dependent variations.
To bridge the spectral-granularity gap between HSI and MSI, we introduce SKT which reparameterizes each MSI embedding kernel as a learnable combination of fine-grained HSI-pretrained kernel slices, thereby transferring spectral priors in parameter space while adapting them to broad-band MSI observations.
Experiments on four HSI and MSI benchmarks demonstrate leading performance on three benchmarks and competitive results on EuroSAT. Ablation studies further validate the effectiveness of each component.
The main contributions are summarized as follows:
\begin{itemize}
    \item We formulate heterogeneous spectral foundation modeling from a shared continuous-spectrum perspective, providing a unified view of diverse band configurations.

    \item We introduce wavelength-aware 3D RoPE to encode physical wavelength and spatial coordinates, enabling fine-grained modeling within a unified representation space.

    \item We propose SKT to transfer pretrained knowledge across spectral granularities in parameter space by constructing MSI embedding kernels from learnable combinations of HSI-pretrained spectral kernel slices.

    \item We construct MultiHyper-200K, a large-scale multi-sensor HSI pretraining dataset, and develop auxiliary-token-assisted MIM to improve cross-sensor representation learning.

\end{itemize}

\section{Related work}

\paragraph{Spectral Foundation Models.}
Spectral foundation models have demonstrated strong performance in remote sensing interpretation~\cite{huang2025survey}. Existing methods mainly model hyperspectral and multispectral imagery separately. For example, SatMAE~\cite{cong2022satmae} and SpectralGPT~\cite{hong2024spectralgpt} focus on multispectral imagery, whereas SpectralEarth~\cite{braham2025spectralearth}, HyperSIGMA~\cite{wang2025hypersigma}, HyperFree~\cite{li2025hyperfree}, HyperSL~\cite{kong2025hypersl}, and SpecAware~\cite{ji2026specaware} are designed for hyperspectral imagery. More recently, sensor-agnostic methods such as SMARTIES and CARL have attempted to unify heterogeneous spectral observations. However, they typically compress spectral information into fixed-dimensional representations at an early stage, which may discard fine-grained spectral structures and hinder parameter transfer across different band configurations.

\paragraph{Masked Image Modeling.}
MIM is a generative self-supervised learning paradigm that masks a large portion of input patches and reconstructs the missing pixels or features from the visible context, thereby learning transferable representations without manual annotations~\cite{he2022masked, bao2022beit}. 
In remote sensing, MIM has been adapted to domain-specific spatial, spectral, temporal, and scale characteristics. Representative methods such as RingMo~\cite{sun2022ringmo}, ScaleMAE~\cite{reed2023scalemae} and SelectiveMae~\cite{wang2025selectivemae}. These methods have achieved strong transfer performance across classification and segmentation, establishing MIM as a fundamental pretraining paradigm for remote sensing representation learning.

\paragraph{Principle of Spectral Imaging.}

Spectral imaging measures an underlying continuous spectrum through a set of discrete spectral intervals~\cite{wei2015hyperspectral}. Let $\mathcal{X}_{p}(\lambda)$ denote the continuous spectral signal at spatial position $p$. The observation of the $i$-th band acquired by sensor $s$ is formulated as
\begin{equation}
\mathcal{X}_{p}^{s}(i)
=
\int_{\Omega_i^{s}}
\mathcal{X}_{p}(\lambda)
r_i^{s}(\lambda)\,d\lambda,
\label{eq:spectral_imaging}
\end{equation}
where $\Omega_i^{s}$ and $r_i^{s}(\lambda)$ denote the spectral sampling interval and response function, respectively. Therefore, spectral images acquired by different sensors can be regarded as discrete observations of the same continuous spectrum under different spectral sampling configurations. Their spectral heterogeneity mainly arises from variations in the number, location, and width of the sampling intervals.

\paragraph{Rotary Positional Encoding.}
RoPE~\cite{su2024roformer} encodes relative positions through coordinate-dependent rotations of query and key features. It has been extended from 1D sequences to 2D visual coordinates and adopted in vision foundation models, such as EVA-02\cite{fang2024eva} and DINOv3~\cite{simeoni2025dinov3}. Spectral images naturally involve spatial dimensions  $(h,w)$ and one spectral dimension $\lambda$, motivating a 3D RoPE over $(\lambda,h,w)$. By using physical wavelength instead of discrete channel indices, heterogeneous HSI and MSI bands can be represented within a consistent spatial-spectral coordinate system.

\begin{figure*}[t]
\centering
\includegraphics[width=0.90\textwidth]{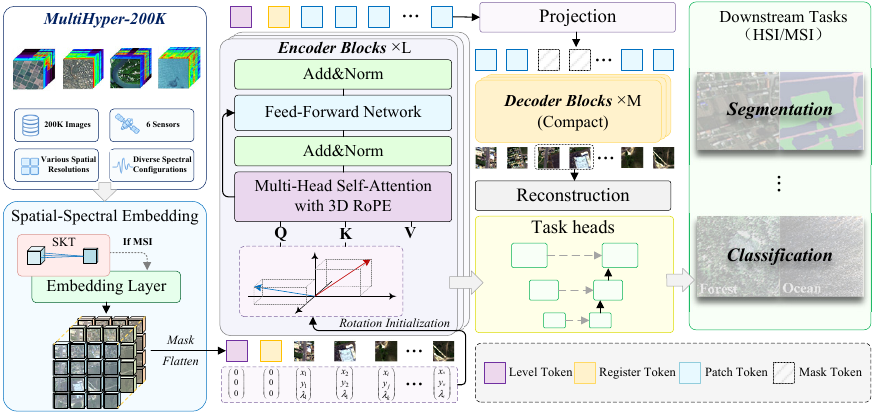} % Reduce the figure size so that it is slightly narrower than the column.
\caption{Overview of SpecUltra. Heterogeneous spectral images are tokenized along the spatial and spectral dimensions and encoded within a unified spatial-spectral representation space using wavelength-aware 3D RoPE. SpecUltra is pretrained on MultiHyper-200K with auxiliary-token-assisted MIM to accommodate radiometric, sensor, and acquisition variations. For MSI adaptation, SKT constructs MSI-compatible embedding kernels from learnable combinations of HSI-pretrained spectral kernel slices, enabling parameter-space transfer across spectral granularities.}
\label{fig:main}
\end{figure*}

\section{Method}
\subsection{Overview}

As illustrated in Fig.~\ref{fig:main}, SpecUltra models heterogeneous spectral images within a unified spatial-spectral representation space. It tokenizes each input with 3D convolution and encodes relative wavelength and spatial coordinates using wavelength-aware 3D RoPE. SpecUltra is pretrained on MultiHyper-200K with auxiliary-token-assisted MIM to accommodate radiometric and sensor-dependent variations. For MSI adaptation, SKT transfers HSI-pretrained embedding parameters across spectral granularities by aggregating fine-grained spectral kernel slices into MSI kernels.

\subsection{Unified Spatial-Spectral Embedding}

Given an input HSI $\mathbf{X}\in\mathbb{R}^{C\times H\times W}$ with a variable number of spectral bands $C$, we first employ a learnable 3D convolutional kernel
$\mathbf{W}^{\mathrm H}\in\mathbb{R}^{D\times1\times K\times S\times S}$
to construct fine-grained spatial-spectral tokens, where $D$ is the embedding dimension, $K$ is the spectral patch size, and $S\times S$ is the spatial patch size. The operation is formulated as:
\begin{equation}
\hat{\mathbf{X}}
=
\operatorname{Conv3D}_{(K,S,S)}
\left(\mathbf{X};\mathbf{W}^{\mathrm H}\right)
\in
\mathbb{R}^{D\times C_K\times H_S\times W_S},
\end{equation}
where the kernel size and stride are both set to $(K,S,S)$. Here,
$C_K=\left\lfloor C/K\right\rfloor$,
$H_S=\left\lfloor H/S\right\rfloor$, and
$W_S=\left\lfloor W/S\right\rfloor$
denote the spectral depth, height, and width of the embedded feature
$\hat{\mathbf{X}}$, respectively.
The resulting feature volume is then flattened along the spectral and spatial dimensions to obtain the spatial-spectral tokens $\mathbf{T}$:
\begin{equation}
\mathbf{\hat{X}}
\xrightarrow{\operatorname{flatten}}
\mathbf{T}=\{\mathbf{t}_1,\mathbf{t}_2,...,\mathbf{t}_N\}\in\mathbb{R}^{N\times D},
\end{equation}
where $N=C_K\times H_S\times W_S$ is the number of output tokens and $D$ is the embedding dimension.

Then, we assign each token a spatial-spectral coordinate based on its wavelength and spatial location.
Each token $\mathbf{t}_i$ is associated with a normalized coordinate
$\mathbf{p}_i=(\hat{\lambda_i}, \hat{h_i}, \hat{w_i})\in[-1,1]^3$, and each $\mathbf{p}_i$ can be computed as follows:
\begin{equation}
\hat{\lambda_i} = \frac{2\lambda_i}{2500}-1, 
\hat{h_i} = \frac{2h_i}{H_S}-1,
\hat{w_i} = \frac{2w_i}{W_S}-1,
\end{equation}
where $\lambda_i$ and $(h_i,w_i)$ denote the mean wavelength and spatial token-grid index of token $\mathbf{t}_i$ , respectively. The constant $2500$ nm is used as a common upper wavelength bound for typical HSIs.

Based on the normalized coordinates, we parameterize 3D RoPE to model spatial-spectral tokens within a unified coordinate system through self-attention. For multi-head attention with $N_{\mathrm{h}}$ heads, each head has dimension $D_h=D/N_{\mathrm{h}}$. Since the rotary dimensions are equally divided among the spectral, height, and width axes, with each rotation operating on a channel pair, we require $D_h\equiv0\pmod{6}$. The rotary periods are geometrically spaced as:
\begin{equation}
P_m=\alpha^{\frac{2m}{D_h/3}},
\quad m=1,2,\ldots,M,
\end{equation}
where $M=D_h/6$ is the number of rotary channel pairs assigned to each axis, and $\alpha$ is set to 10000 following the original RoPE.
The rotary angles are computed given the normalized coordinate $\mathbf{p}_i$ as follows:
\begin{equation}
\theta_{i,m}^{(\lambda)}=\frac{2\pi \hat \lambda_i}{P_m},
\theta_{i,m}^{(h)}=\frac{2\pi \hat h_i}{P_m},
\theta_{i,m}^{(w)}=\frac{2\pi \hat w_i}{P_m},
\end{equation}
where $P_m$ denotes the rotary period at frequency index $m$.
Subsequently, the rotary angles are concatenated and duplicated to construct the rotation vector $\boldsymbol{\theta}_i$ for paired-channel rotation:
\begin{equation}
\bar{\boldsymbol{\theta}}_i=
\left[
\theta_{i,1}^{(\lambda)},\ldots,
\theta_{i,1}^{(h)},\ldots,
\theta_{i,1}^{(w)},\ldots,\theta_{i,M}^{(w)}
\right]\in\mathbb{R}^{\frac{D_h}{2}},
\end{equation}

\begin{equation}
\boldsymbol{\theta}_i=
[\bar{\boldsymbol{\theta}}_i,\bar{\boldsymbol{\theta}}_i] \in\mathbb{R}^{D_h}.
\end{equation}

For convenience, we define a rotation operator $\mathrm{rot}(\cdot)$:
\begin{equation}
\operatorname{rot}(\mathbf{x})
=
[-\mathbf{x}^{(2)},\mathbf{x}^{(1)}],
\quad
\mathbf{x}=[\mathbf{x}^{(1)},\mathbf{x}^{(2)}],
\end{equation}
where $\mathbf{x}^{(1)},\mathbf{x}^{(2)}\in\mathbb{R}^{D_h/2}$.
For each attention head, let 
\(\mathbf{q}_i,\mathbf{k}_i\in\mathbb{R}^{D_h}\) denote the query and key vectors of token \(\mathbf{t}_i\). The rotated query and key are computed as follows:
\begin{equation}
\tilde{\mathbf{q}}_i
=
\mathbf{q}_i\odot \cos\boldsymbol{\theta}_i
+
\mathrm{rot}(\mathbf{q}_i)\odot \sin\boldsymbol{\theta}_i,
\end{equation}
\begin{equation}
\tilde{\mathbf{k}}_i
=
\mathbf{k}_i\odot \cos\boldsymbol{\theta}_i
+
\mathrm{rot}(\mathbf{k}_i)\odot \sin\boldsymbol{\theta}_i,
\end{equation}
where $\odot$ denotes element-wise multiplication.
From a geometric perspective, this operation induces an independent planar rotation within each paired-channel subspace. 
Specifically, for the \(k\)-th channel pair 
\((x_{i,k},x_{i,k+D_h/2})\), with the corresponding angle 
\(\phi_{i,k}=(\bar{\boldsymbol{\theta}}_i)_k\), the transformation can be written as:
\begin{equation}
\begin{bmatrix}
x'_{i,k}\\
x'_{i,k+D_h/2}
\end{bmatrix}
=
\begin{bmatrix}
\cos\phi_{i,k} & -\sin\phi_{i,k}\\
\sin\phi_{i,k} & \cos\phi_{i,k}
\end{bmatrix}
\begin{bmatrix}
x_{i,k}\\
x_{i,k+D_h/2}
\end{bmatrix}.
\end{equation}
Accordingly, the proposed 3D RoPE performs rotary modulation by assigning separate channel groups to the wavelength, height, and width coordinates, enabling self-attention to model relative spatial-spectral geometry and align tokens from heterogeneous sensors within a unified coordinate system.

\subsection{Spectral Kernel Transformation}

The large spectral-granularity gap between HSI and MSI results in incompatible embedding parameterizations. Directly replacing the HSI embedding layer discards pretrained spectral priors and disrupts its compatibility with the backbone. We therefore formulate MSI adaptation as a parameter-space transformation, constraining the MSI embedding kernel to the space induced by HSI-pretrained spectral kernel slices.

Given the HSI embedding kernel
$\mathbf{W}^{\mathrm H}\in\mathbb{R}^{D\times1\times K\times S\times S}$,
its $k$-th spectral slice for output channel $d$ is defined as
\begin{equation}
\mathbf{W}^{\mathrm H}_{d,k}
=
\mathbf{W}^{\mathrm H}(d,1,k,:,:)
\in\mathbb{R}^{S\times S}.
\end{equation}
The shared MSI embedding kernel
$\mathbf{W}^{\mathrm M}\in\mathbb{R}^{D\times1\times1\times S\times S}$
is parameterized as
\begin{equation}
\mathbf{W}^{\mathrm M}_{d}
=
\sum_{k=1}^{K}
A_{d,k}\mathbf{W}^{\mathrm H}_{d,k},
\quad
A_{d,k}
=
\frac{\exp(\widetilde{A}_{d,k})}
{\sum_{j=1}^{K}\exp(\widetilde{A}_{d,j})},
\label{eq:skt_kernel}
\end{equation}
where
$\widetilde{\mathbf{A}}\in\mathbb{R}^{D\times K}$
contains learnable parameters and
$\mathbf{A}\in\mathbb{R}^{D\times K}$
is the normalized transformation matrix.
This transformation transfers the spectral aggregation principle from observation space to parameter space. Since
$A_{d,k}\geq0$
and
$\sum_{k=1}^{K}A_{d,k}=1$,
each MSI kernel lies in the convex hull of the corresponding HSI-pretrained slices. SKT thus preserves pretrained spectral priors while learning an MSI-compatible embedding parameterization.

For an MSI input $\mathbf{X}^{\mathrm M}\in\mathbb{R}^{C_{\mathrm M}\times H\times W}$, $\mathbf{W}^{\mathrm M}$ is applied independently to each band through convolution operation, producing:
\begin{equation}
\hat{\mathbf{X}}^{\mathrm M}\in
\mathbb{R}^{D\times C_{\mathrm M}\times H_S\times W_S}.
\end{equation}
The resulting feature volume retains explicit spectral and spatial axes, allowing each token to be indexed by its wavelength and spatial coordinates. The transformed MSI features therefore remain compatible with the pretrained backbone and can be directly processed using wavelength-aware 3D RoPE.

\subsection{Auxiliary-Tokens-Assisted MIM}

Multi-source HSI data share transferable spatial-spectral structures while exhibiting substantial variations in sensors, acquisition conditions, and radiometric products. To improve representation learning under such heterogeneity, we introduce two types of auxiliary tokens during masked image modeling: register tokens and a radiometric-level token.

The register tokens
$\mathbf{R}\in\mathbb{R}^{N_r\times D}$
provide additional latent capacity for capturing sensor- and acquisition-dependent variations, thereby encouraging patch tokens to focus on intrinsic and transferable spatial-spectral patterns. The level token specifies the radiometric form of the input and is selected from a learnable embedding table:
\begin{equation}
g_{\ell}
=
\mathbf{E}_{\mathrm{level}}[\ell],
\qquad
\mathbf{E}_{\mathrm{level}}
\in
\mathbb{R}^{2\times D},
\end{equation}
where $\ell\in\{0,1\}$ denotes radiance or reflectance data.

Given the visible patch tokens $\mathbf{T}_{\mathrm{vis}}$, the encoder input is constructed as
\begin{equation}
\mathbf{Z}_{\mathrm{enc}}
=
\mathcal{E}
\left(
[g_{\ell},\mathbf{R},\mathbf{T}_{\mathrm{vis}}]
\right).
\end{equation}
The decoder then reconstructs the masked patches using learnable mask tokens:
\begin{equation}
\hat{\mathbf{X}}
=
\mathcal{D}
\left(
[\mathbf{Z}_{\mathrm{enc}},\mathbf{T}_{\mathrm{mask}}]
\right).
\end{equation}
Both auxiliary tokens are used only to condition the encoder representations and are excluded from the reconstruction decoder and downstream task heads.

Following ~\cite{ji2026specaware}, we optimize the masked prediction task using a hybrid objective that jointly accounts for radiometric fidelity and spectral consistency. Specifically, the Charbonnier loss mitigates the influence of noise and large reconstruction errors, while the spectral angle mapper (SAM) loss constrains the angular discrepancy between the reconstructed and target spectra, thereby preserving spectral structures under magnitude variations. The final pretraining objective is defined as
\begin{equation}
\begin{aligned}
\mathcal{L}
&=
\alpha\mathcal{L}_{\mathrm{char}}
+
\beta\mathcal{L}_{\mathrm{sam}}
\\
&=
\frac{\alpha}{|\mathcal{M}|}
\sum_{i\in\mathcal{M}}
\sqrt{
\left\|
\mathbf{x}_{i}-\hat{\mathbf{x}}_{i}
\right\|_{2}^{2}
+\epsilon^{2}
}
\\
&+
\frac{\beta}{|\mathcal{M}|}
\sum_{i\in\mathcal{M}}
\left(
1-
\frac{
\mathbf{x}_{i}^{\top}\hat{\mathbf{x}}_{i}
}{
\left\|\mathbf{x}_{i}\right\|_{2}
\left\|\hat{\mathbf{x}}_{i}\right\|_{2}
+\epsilon
}
\right).
\end{aligned}
\label{eq:pretraining_loss}
\end{equation}
Where, $\mathcal{M}$ denotes the set of masked patches, and
$\mathbf{x}_{i}$ and $\hat{\mathbf{x}}_{i}$ are the corresponding target and reconstructed spectral vectors, $\alpha$ and $\beta$ are set to 1 by default, $\epsilon$ ensures numerical stability. 

\section{Experiments}

\subsection{Dataset Description}

\paragraph{MultiHyper-200K:}
To facilitate large-scale multi-sensor pretraining of SpecUltra, we constructed a HSI dataset from six sensors: AVIRIS, AVIRIS-3, DESIS, EnMAP, GF-5, and HyspIRI. The collected imagery covers diverse geographic regions worldwide.
After manual quality inspection, the remaining images were cropped into approximately 200K non-overlapping $256\times256$ samples for pretraining. The details of MultiHyper-200K are summarized in Table~\ref{table:dataset}.

\begin{table}[h]
\centering

\resizebox{\columnwidth}{!}{
\begin{tabular}{lcccccc}
\toprule
Sensors 
& Spec. Range 
& Spec. Res
& Spat. Res 
& Bands
& Data
& Numbers \\
& (nm) 
& (nm) 
& (m) 
& (\#) 
& Level
& (\#)\\
\midrule
AVIRIS   & 380-2500 & 10        & 0.5-20 & 224 & L1/L2  & 94,207 \\
AVIRIS-3 & 390-2500 & 7.4       & 0.3-10 & 224 & L2     & 29,866 \\
HyspIRI  & 380-2500 & 10        & 0.3-17 & 224 & L2     & 12,012 \\
EnMAP    & 420-2450 & 6.5/10    & 30      & 246 & L2     & 23,207 \\
DESIS    & 402-1000 & 2.55      & 30      & 235 & L2     & 27,517 \\
GF-5     & 400-2500 & 5/10      & 30      & 330 & L1     & 13,689 \\
\midrule
\textbf{Total} &  &  &  &  &   & \textbf{200,498} \\
\bottomrule
\end{tabular}
}
\caption{Main characteristics of the pretraining datasets.}
\label{table:dataset}
\end{table}

\paragraph{Downstream datasets:}
To evaluate SpecUltra across heterogeneous sensors, spectral modalities, and downstream tasks, we adopt four representative HSI and MSI benchmarks covering airborne and satellite platforms, semantic segmentation, and scene classification.
For semantic segmentation, Qingpu~\cite{ji2025patchout} is an AMMIS HSI of size $20{,}480\times2{,}944$ with 251 bands and 20 classes. Following~\cite{ji2025patchout}, spatially disjoint regions are used for training and testing to avoid spatial leakage. SegMunich~\cite{hong2024spectralgpt} is a Sentinel-2 MSI of size $3{,}847\times2{,}958$ with 10 bands and 13 classes, evaluated using the SpectralGPT protocol. These datasets assess pixel-level transfer across airborne HSI and satellite MSI.
For scene classification, HRSSC~\cite{li2025ssfnet} contains 1,445 AVIRIS-C HSI patches of size $256\times256$, with 224 bands and 11 classes. We use 10\% of the samples for training and the remainder for testing. EuroSAT~\cite{helber2019eurosat} contains 27,000 Sentinel-2 MSI patches of size $64\times64$, with 13 bands and 10 classes. We use 100 samples per class for training and the rest for testing. These datasets evaluate image-level generalization across spectral granularities under limited supervision.

\subsection{Implementation details}

SpecUltra was pretrained with Auxiliary-Token-Assisted MIM using a masking ratio of 0.75 following the common practice in MAE~\cite{he2022masked}. The embedding kernel and stride were set to $(10,16,16)$ along the spectral and spatial dimensions. Reflectance products retained their original physical scale, while radiance products were empirically scaled during preprocessing. A learnable level token indicated the radiometric product type, and five register tokens were introduced to capture sensor-specific variations. We trained SpecUltra-S and SpecUltra-B for 200 epochs using AdamW with cosine learning-rate decay, an initial learning rate of $10^{-3}$, an effective batch size of 256, a weight decay of 0.05, and $\beta_1=0.9$, $\beta_2=0.95$. Mixed-precision training and random horizontal and vertical flips were adopted. The model configurations are summarized in Table~\ref{tab:model_configs}.

During fine-tuning, all models were trained for 100 epochs with a backbone learning rate of 
$1\times10^{-4}$. The spatial patch size was set to 16 for classification and 8 for segmentation, where the pretrained embedding kernels were spatially downsampled to match the smaller patch size. The batch sizes were set to 16, 16, 32, and 64 for Qingpu, HRSSC, SegMunich, and EuroSAT, respectively. The same fine-tuning protocol was applied to all comparison methods to ensure a fair evaluation.
\begin{table}[t]
\centering
\renewcommand{\arraystretch}{1.05}
\setlength{\tabcolsep}{6pt}
\begin{tabular}{lcccc}
\toprule
Model
& Dim.
& Depth
& Heads
& MLP
 \\
\midrule
SpecUltra-S & 384/192 & 12/8 & 4/4 & 4.0 \\
SpecUltra-B  & 576/192 & 12/8 & 8/8 & 4.0\\
\bottomrule
\end{tabular}
\caption{Configurations of SpecUltra variants (Enc./Dec.).}
\label{tab:model_configs}
\end{table}

\begin{table*}[t]
\centering
\renewcommand{\arraystretch}{1.0}
\setlength{\tabcolsep}{3pt}

\begin{tabular}{llcccc}
\toprule

\rowcolor{gray!15}
&
& \multicolumn{2}{c}{\textbf{Segmentation (OA/F1/mIoU)}}
& \multicolumn{2}{c}{\textbf{Classification (OA/F1/Kappa)}} \\

\rowcolor{gray!15}
\multirow[c]{-2}{*}[-0.6ex]{\textbf{Method}}
& \multirow[c]{-2}{*}[-0.6ex]{\textbf{Pretraining Data}}
& \textbf{Qingpu (HSI)}
& \textbf{SegMunich (MSI)}
& \textbf{HRSSC (HSI)}
& \textbf{EuroSAT (MSI)} \\

\midrule

SatMAE
& fMoW-S2
& 91.57/46.81/36.93
& \second{82.29}/\second{63.40}/\second{50.21}
& 79.01/66.53/75.61
& \second{94.02}/\second{93.80}/\second{93.34}
\\

SpectralGPT
& BigEarthNet
& 91.62/47.32/37.65
& 81.63/61.27/49.21
& 82.67/70.26/79.27
& \best{94.69}/\best{94.49}/\best{94.48}
\\

\midrule

SpectralEarth
& SpectralEarth
& 92.23/49.74/40.26
& 82.03/62.98/49.96
& 82.41/73.79/79.58
& 92.32/83.70/91.46
\\

HyperSigma
& HyperGlobal-450K
& 91.96/46.99/38.47
& 80.37/58.20/46.81
& \second{83.02}/\best{74.63}/\second{80.34}
& 91.54/82.89/90.59
\\

HyperFree
& HyperSeg
& 90.85/44.27/35.62
& 81.18/60.56/47.66
& 80.48/69.50/77.47
& 91.19/82.52/90.20
\\

SpecAware
& Hyper-400K
& 93.38/50.38/41.18
& 81.50/61.72/48.70
& 82.64/69.55/79.78
& 90.13/89.70/89.02
\\
\midrule
SMARTIES
& fMoW-S2
& \second{93.51}/\second{55.93}/\second{44.73}
& 81.33/61.32/48.26 
& 81.86/69.91/78.89
& 91.27/90.92/90.29
\\

CARL
& HySpecNet
& 92.20/49.28/39.50  
& 80.05/58.80/45.95  
& 79.86/69.81/76.77
& 90.28/89.93/89.19
\\

\midrule

\rowcolor{blue!4}
\textbf{SpecUltra-B}
& MultiHyper-200K
& \best{94.69}/\best{56.87}/\best{47.04}
& \best{82.74}/\best{64.07}/\best{50.90}
& \best{85.73}/\second{74.14}/\best{83.42}
& 93.74/93.45/93.03
\\

\bottomrule
\end{tabular}

\caption{Comparison with SOTA methods on HSI and MSI downstream
benchmarks(\best{best}/\second{second}).(\%).}
\label{tab:main_downstream}
\end{table*}

\begin{figure*}[!h]
\centering
\includegraphics[width=0.9\textwidth]{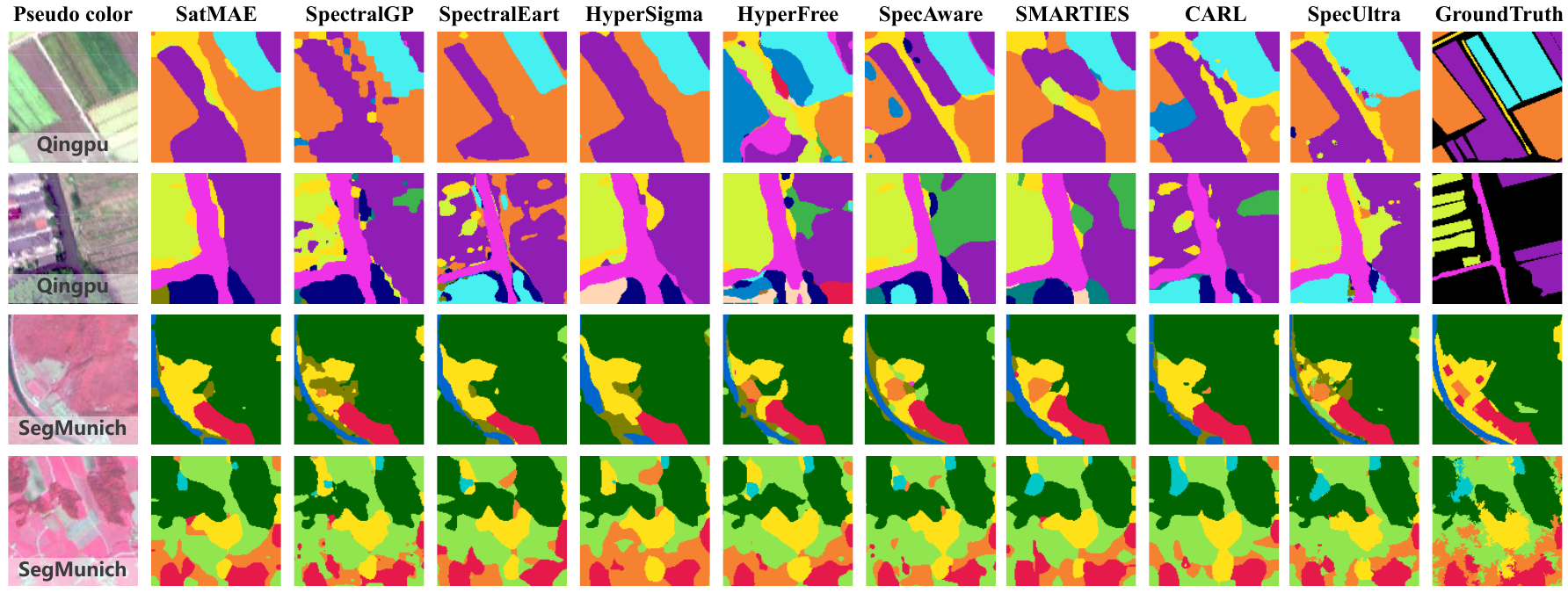} % Reduce the figure size so that it is slightly narrower than the column.
\caption{Qualitative comparison of segmentation results on Qingpu and SegMunich datasets.}
\label{fig:seg_vis}
\end{figure*}

\subsection{Main Results}

\paragraph{Main Comparison.}
We evaluate SpecUltra on two classification benchmarks, EuroSAT and HRSSC, and two segmentation benchmarks, SegMunich and Qingpu, covering both MSI and HSI.
We compare SpecUltra with SMARTIES, CARL, and six representative remote-sensing pretraining methods: SatMAE, SpectralGPT, SpectralEarth, HyperSIGMA, HyperFree, and SpecAware.
For methods incompatible with heterogeneous inputs, a prefix adapter is used for channel alignment. 
We report OA, F1, and mIoU for segmentation, and OA, F1, and Kappa for classification. Results are averaged over five runs with different random seeds.

\paragraph{Finetuning.}

As summarized in Table~\ref{tab:main_downstream}, SpecUltra achieves strong performance across HSI and MSI segmentation and classification benchmarks. Under the spatially disjoint protocol on Qingpu, it obtains 94.69\% OA, 56.87\% F1, and 47.04\% mIoU, surpassing the strongest baseline by 1.18, 0.94, and 2.31 points, respectively, indicating robust spatial transfer without relying on neighboring-region leakage. On SegMunich, it further improves the best competing results by 0.45, 0.67, and 0.69 points, demonstrating effective adaptation from HSI pretraining to sparse MSI bands.
For classification, SpecUltra achieves the best OA and Kappa on HRSSC, outperforming HyperSigma by 2.71 and 3.08 points, while obtaining the second-best F1 with a gap of only 0.49 points. Given the limited training split, these results reflect strong sample-efficient transfer. On EuroSAT, it remains competitive with MSI-specific models and consistently outperforms all HSI-pretrained and sensor-agnostic methods, showing that SKT and wavelength-aware modeling alleviate the HSI-MSI spectral-granularity gap. The qualitative results in Fig.~\ref{fig:seg_vis} further show more coherent regions, clearer boundaries, and fewer fragmented predictions. Overall, SpecUltra generalizes effectively across heterogeneous sensors, spectral granularities, and downstream tasks.

\paragraph{Backbone Probing.}

To fairly evaluate representation transfer across heterogeneous spectral configurations, we freeze the pretrained backbones and optimize only the adaptation modules and linear classifiers for all methods. As shown in Table~\ref{tab:linear_probe}, the proposed method achieves the best performance on HRSSC, surpassing the strongest baselines by 0.33 OA, 1.31 F1, and 1.85 Kappa points. On EuroSAT, it trails the MSI-specific methods by at most 0.98 points, while outperforming all HSI-pretrained methods across the three metrics. Compared with sensor-agnostic methods, it achieves the highest OA and Kappa while remaining competitive in F1. These results demonstrate the strong transferability of SpecUltra across heterogeneous HSI and MSI configurations

\begin{table}[!h]
\centering
\renewcommand{\arraystretch}{0.90}
\setlength{\tabcolsep}{3pt}

\begin{tabular}{lcccccc}
\toprule
\multirow{2}{*}{\textbf{Method}}
& \multicolumn{3}{c}{HRSSC}
& \multicolumn{3}{c}{EuroSAT} \\
\cmidrule(lr){2-4}\cmidrule(lr){5-7}
& OA
& F1
& Kappa
& OA
& F1
& Kappa \\
\midrule

SatMAE
& 74.92 & 57.91 & 70.56  
& \second{83.52} & \best{82.34} & \second{81.66} \\   

SpectralGPT
& 76.38 & 58.46 & 70.99 
& \best{83.58} & \second{82.27} & \best{81.73} \\ 

\midrule

SpectralEarth
& 77.47 & 68.41 & 73.82
& 82.57 & 74.10 & 80.59 \\

HyperSigma
& 80.17 & \second{69.63} & 76.99
& 81.66 & 73.47 & 79.59 \\ 

HyperFree
& 80.63 & 68.32 & 77.39  
& 77.45 & 69.27 & 74.92 \\  

SpecAware
& \second{81.85}  & 68.45 & \second{78.88} 
& 80.50  & 80.00 & 78.30 \\  

\midrule

SMARTIES
& 78.94 & 65.60 & 75.45  
& 82.87 & 82.35 & 80.93 \\  

CARL
& 76.08  & 63.03 & 72.12   
& 82.59  & 81.60 & 80.60 \\   

\midrule

SpecUltra-B
& \best{82.18} & \best{70.94} & \best{80.73}
& 82.93 & 81.37 & 80.94\\ 

\bottomrule
\end{tabular}%
\caption{Backbone probing results on classification benchmarks.}
\label{tab:linear_probe}
\end{table}

\paragraph{Model scalability.}

\begin{figure}[h]
\centering
\includegraphics[width=\columnwidth]{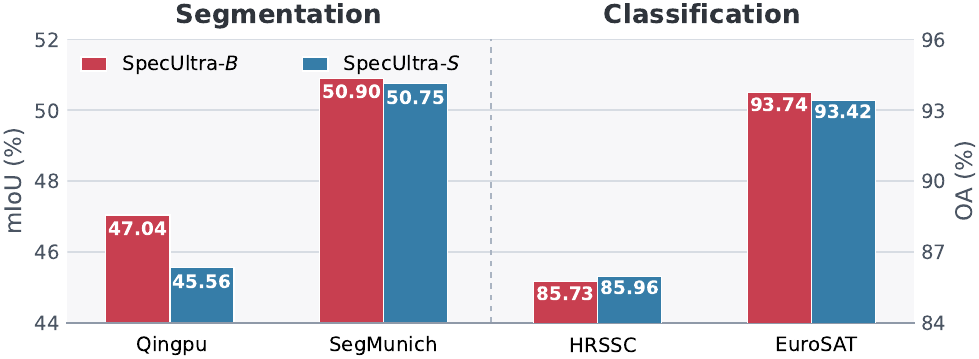}
\caption{Comparison of SpecUltra-S and SpecUltra-B.}
\label{fig:model_scalability}
\end{figure}

As shown in Fig.~\ref{fig:model_scalability}, scaling SpecUltra from small to base improves mIoU by 1.48 and 0.15 points on Qingpu and SegMunich, respectively, and increases accuracy by 0.32 points on EuroSAT.
On HRSSC, the Base variant performs slightly worse by 0.23 points, likely because its limited training set restricts the effective optimization of the larger model. Overall, increased model capacity generally improves downstream performance, particularly for dense prediction tasks with sufficient training data.

\begin{table*}[t]
\centering
\renewcommand{\arraystretch}{1.1}
\setlength{\tabcolsep}{4pt}
\resizebox{\textwidth}{!}{
\begin{tabular}{lcccccccc|cc}
\toprule
\textbf{Metric}
& \textbf{SatMAE}
& \textbf{SpectralGPT}
& \textbf{SpectralEarth}
& \textbf{HyperSigma}
& \textbf{HyperFree}
& \textbf{SpecAware}
& \textbf{SMARTIES}
& \textbf{CARL}
& \textbf{SpecUltra-S}
& \textbf{SpecUltra-B}
\\
\midrule
Params
& 86.09 M
& 85.41 M
& 236.19 M
& 236.19 M
& 151.14 M
& 89.42 M
& 88.62 M
& 71.93 M
& 24.69 M
& 55.31 M
\\

FLOPs
& 523.70 G
& 212.97 G
& 229.44 G
& 229.44 G
& 57.96 G
& 45.16 G
& 3313.01 G
& 63.02 G
& 324.65 G
& 727.59 G
\\

\bottomrule
\end{tabular}
}

\caption{Comparison of model complexity.}
\label{tab:complexity}
\end{table*}

\paragraph{Computational Efficiency.}

As shown in Table~\ref{tab:complexity}, SpecUltra uses fewer parameters than most competing models. Its higher FLOPs arise from jointly tokenizing the spectral and spatial dimensions, which produces more tokens for fine-grained spatial-spectral modeling. Thus, the additional computation mainly reflects denser token interactions.

\subsection{Ablation Studies}
We conduct controlled ablation studies to evaluate the individual contributions of wavelength-aware 3D RoPE, SKT, and auxiliary-token-assisted pretraining. To ensure a fair comparison while reducing computational cost, all variants adopt the tiny model and are pretrained for 30 epochs under identical settings. Performance is evaluated using mIoU for segmentation tasks and OA for scene classification tasks, with all ablation results averaged over five runs.

\paragraph{Effect of Wavelength-Aware 3D RoPE.}

Table~\ref{tab:rope_ablation} compares different positional encodings. Replacing band indices with wavelengths using Wave PE \& 2D RoPE improves Index PE by 0.89, 2.14, 1.22, and 2.31 points on the four datasets, respectively. Joint wavelength--spatial encoding further gains 1.59, 0.13, 0.86, and 0.76 points, achieving the best overall performance and demonstrating the transferability of unified 3D spatial--spectral coordinates.

\paragraph{Effect of SKT.}

Table~\ref{tab:SKT_ablation} compares different MSI embedding strategies. Prefix Adapter provides limited gains and slightly degrades SegMunich performance, whereas Mean Aggregation improves over random initialization by 0.22 mIoU and 1.75 OA, confirming the benefit of reusing HSI-pretrained kernels. SKT achieves the best results, further surpassing Mean Aggregation by 0.40 mIoU on SegMunich and 1.68 OA on EuroSAT. These gains show that learnable kernel transformation better bridges the spectral-granularity gap between HSI and MSI.

\paragraph{Effect of Register Tokens.}

\begin{figure}[!h]
\centering
\includegraphics[width=\columnwidth]{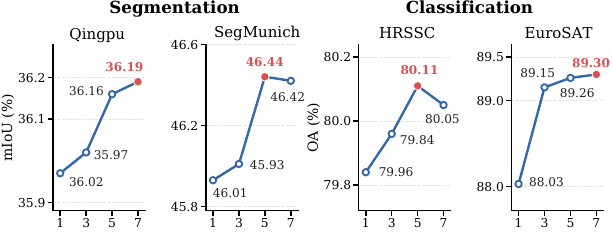} 
\caption{Effect of the number of register tokens.}
\label{fig:reg_tokens}
\end{figure}

Table~\ref{tab:register_ablation} shows that both level and register tokens consistently improve performance. Their combination achieves the best results, surpassing the token-free baseline by 1.27, 2.06, 1.17, and 2.00 points on Qingpu, SegMunich, HRSSC, and EuroSAT, respectively. This suggests that level tokens provide radiometric-level cues, while register tokens capture sensor- and acquisition-dependent variations. As shown in Fig.~\ref{fig:reg_tokens}, performance generally saturates at five register tokens; increasing the number to seven yields only marginal gains on Qingpu and EuroSAT while slightly degrading performance on SegMunich and HRSSC. Therefore, we use five register tokens by default.

\begin{table}[h]
\centering
\renewcommand{\arraystretch}{1.0}
\setlength{\tabcolsep}{2pt}
\resizebox{\columnwidth}{!}{
\begin{tabular}{lcccc}
\toprule
Position Encoding
& \shortstack{Qingpu}
& \shortstack{SegMunich}
& \shortstack{HRSSC}
& \shortstack{EuroSAT}\\
\midrule
Index PE
& 33.68$\pm$0.76 
& 44.17$\pm$0.24
& 78.03$\pm$1.43 
& 86.19$\pm$1.90
\\

Wave PE \& 2D RoPE
& 34.57$\pm$0.33 
& 46.31$\pm$0.19
& 79.25$\pm$1.22 
& 88.50$\pm$1.30 \\

Wave 3D RoPE
& 36.16$\pm$0.50 
& 46.44$\pm$0.20
& 80.11$\pm$1.68 
& 89.26$\pm$1.22 \\
\bottomrule
\end{tabular}%
}
\caption{Effect of wavelength-aware 3D RoPE(mean$\pm$std\%).}
\label{tab:rope_ablation}
\end{table}

\begin{table}[h]
\centering
\renewcommand{\arraystretch}{1.0}
\setlength{\tabcolsep}{10pt}
\small
% \resizebox{\columnwidth}{!}{
\begin{tabular}{lcc}
\toprule
MSI Embedding
& SegMunich
& EuroSAT \\
\midrule
Random Initialization
& 50.37$\pm$0.30
& 91.41$\pm$1.72 \\

Prefix Adapter
& 50.12$\pm$0.28 
& 92.06$\pm$1.98\\

Mean Aggregation
& 50.43$\pm$0.21
& 92.96$\pm$1.51\\

Random Aggregation
& 50.50$\pm$0.26
& 93.08$\pm$1.64\\

SKT
& 50.90$\pm$0.25 
& 93.74$\pm$1.76 \\
\bottomrule
\end{tabular}%
\caption{Effect of SKT on MSI transfer on SpecUltra-B.}
\label{tab:SKT_ablation}
\end{table}

\begin{table}[!h]
\centering
\renewcommand{\arraystretch}{1.0}
\setlength{\tabcolsep}{2pt}
\resizebox{\columnwidth}{!}{
\begin{tabular}{lcccc}
\toprule
Token Configuration
& Qingpu
& SegMunich
& HRSSC
& EuroSAT
\\
\midrule
None
& 34.89$\pm$1.97 
& 44.38$\pm$0.29
& 78.94$\pm$1.43 
& 87.26$\pm$2.11 \\

Level Token
& 35.63$\pm$1.05 
& 45.19$\pm$0.32
& 79.16$\pm$1.31 
& 87.83$\pm$2.36 \\

Register Tokens        
& 35.77$\pm$1.27 
& 45.03$\pm$0.28
& 79.50$\pm$1.87
& 88.42$\pm$1.67 \\

Level + Register Tokens 
& 36.16$\pm$0.80 
& 46.44$\pm$0.19
& 80.11$\pm$1.62 
& 89.26$\pm$1.04 \\
\bottomrule
\end{tabular}
}
\caption{Ablation resluts of register and level tokens.}
\label{tab:register_ablation}
\end{table}

\section{Conclusion}
This study develops SpecUltra, a unified foundation model for learning transferable representations from heterogeneous spectral imagery. SpecUltra employs wavelength-aware 3D RoPE to preserve fine-grained spatial-spectral structure across diverse band configurations and is pretrained on MultiHyper-200K with auxiliary-token-assisted MIM to improve cross-sensor representation learning. To bridge the spectral-granularity gap between HSI and MSI, SKT performs parameter-space adaptation by constructing MSI embedding kernels from HSI-pretrained spectral kernel slices. Experiments on classification and segmentation benchmarks demonstrate strong performance across multiple sensors and spectral modalities. These results suggest that SpecUltra provides a practical foundation for unified spectral remote sensing. Future work will explore broader sensor coverage, additional downstream tasks, and more efficient spatial-spectral modeling.

\bibliography{aaai2027}


\end{document}